# CONVERGENCE ANALYSIS OF DIFFERENTIAL EVOLUTION VARIANTS ON UNCONSTRAINED GLOBAL OPTIMIZATION FUNCTIONS

G.Jeyakumar[1]   C.Shanmugavelayutham[2]

[1, 2] Assistant Professor Department of Computer Science and Engineering Amrita School of Engineering, Amrita Vishwa Vidya Peetham, Coimbatore, Tamil Nadu, India

[1]g_jeyakumar@cb.amrita.edu
[2]cs_velayutham@cb.amrita.edu

## ABSTRACT

*In this paper, we present an empirical study on convergence nature of Differential Evolution (DE) variants to solve unconstrained global optimization problems. The aim is to identify the competitive nature of DE variants in solving the problem at their hand and compare. We have chosen fourteen benchmark functions grouped by feature: unimodal and separable, unimodal and nonseparable, multimodal and separable, and multimodal and nonseparable. Fourteen variants of DE were implemented and tested on fourteen benchmark problems for dimensions of 30. The competitiveness of the variants are identified by the Mean Objective Function value, they achieved in 100 runs. The convergence nature of the best and worst performing variants are analyzed by measuring their Convergence Speed ($C_s$) and Quality Measure ($Q_m$).*

## 1. INTRODUCTION

Evolutionary algorithms (EA) have been widely used to solve optimization problems. Differential Evolution [1] is an EA proposed to solve optimization problems, mainly to continuous search spaces. The DE algorithm, a stochastic population-based search method, has been successfully applied to many global optimization problems [2]. As traditional EAs, several optimization problems have been successfully solved by using DE [3]. It shows superior performance in both widely used benchmark functions and real-world application [4, 5]. DE shares similarities with traditional EAs. As in other EAs, two main processes that derive the evolution are the perturbation process (crossover and mutation) which ensures the exploration of the search space and the selection process which ensures the exploitation properties of the algorithm. Both perturbation and the selection process are simpler than those used in other evolutionary algorithms. In the case of DE, the perturbation of a population element is done by probabilistically replacing it with an offspring obtained by adding to a randomly selected element a perturbation proportional with the difference between other two randomly selected elements. The selection is done by one to one competition between the parent and its offspring.

There are three strategy parameters in DE, the population size NP, the crossover rate CR and the scaling factor F. Many works have been done to study the suitable setting of these control parameters [6, 7, 8]. The CR parameter controls the influence of the parent in the generation of the offspring. The F parameter scales the influence of the set of pairs of solutions selected to calculate the mutation value. DE performs the perturbation based on the distribution of the solutions in the current population. In this way, search directions and possible step sizes depend on the location of the individuals selected to calculate the mutation values.

Based on different strategies followed for perturbation, there are various DE variants are exists, they differ in the way how the solution is generated. Besides the suitable setting of control parameters, another important factor when using DE is the selection of the variant. The most popular variant of DE is *rand/1/bin*. There is a nomenclature scheme developed to reference different DE variants. In rand/1/bin, "*DE*" means Differential Evolution, the word "*rand*" indicates that the individuals selected to compute the mutation values are chosen at random, "*1*" is the number of pairs of individuals chosen and finally "*bin*" means that a binomial crossover is used. The algorithm for *rand/1/bin* is presented in Figure. 1.

1. Begin
2. G=0
3. Create random initial population $X_{i,G}$   i =1,…,NP
4. Evaluate  f($X_{i,G}$) i=1,…,NP
5. For G = 1 to MAXGEN Do
6. For i = 1 to NP Do
7.♦ select randomly r1 $\neq$ r2 $\neq$ r3
8. ♦ $j_{rand}$=randint (1,D)
9. ♦ For j=1 to D Do
10. ♦ If(rand$_j$[0,1)<CR or j=$j_{rand}$) Then
11. ♦ $U_{i,j,G+1}$=$X_{r3,j,G}$+F($X_{r1,j,g}$-$X_{r2,j,G}$)
12. ♦ Else
13. ♦ $U_{i,j,G+1}$=$X_{i,j,G}$
14. ♦ End If
15. ♦ End For
16. If  f($U_{i,G+1}$) $\leq$ f($X_{i,G}$) Then
17. $X_{i,G+1}$ = $U_{i,G+1}$
18. Else
19. $X_{i,G+1=}$ $X_{i,G}$
20. End If
21. End For

Figure 1. "*rand/1/bin*" algorithm, steps pointed out with ♦ will change from variant to variant.

With seven  commonly used differential mutation strategies,  as listed in Table 1, and two crossover schemes (binomial and exponential), we get fourteen possible variants of DE. Following the standard DE nomenclature used in the literature, the fourteen DE variants can be written as follows: *rand/1/bin, rand/1/exp, best/1/bin, best/1/exp, rand/2/bin, rand/2/exp, best/2/bin, best/2/exp, current-to-rand/1/bin, current-to-rand/1/exp, current-to-best/1/bin, current-to-best/1/exp, rand-to-best/1/bin and rand-to-best/1/exp*. This paper, an empirical convergence analysis of DE variants has been attempted.

The remainder of the paper is organized as follows. After a brief review of the related work in Section 2, Section 3 details the design of experiments. Section 4 describes the empirical measurements done in our study, Section 5 discusses the simulation results and finally Section 6 concludes the paper.

Table 1. Differential Evolution Variants

| Nomenclature | Variant |
|---|---|
| *rand/1* | $V_{i,G} = X_{r_1^i,G} + F\left(X_{r_2^i,G} - X_{r_3^i,G}\right)$ |
| *best/1* | $V_{i,G} = X_{best,G} + F\left(X_{r_1^i,G} - X_{r_2^i,G}\right)$ |
| *rand/2* | $V_{i,G} = X_{r_1^i,G} + F\left(X_{r_2^i,G} - X_{r_3^i,G} + X_{r_4^i,G} - X_{r_5^i,G}\right)$ |
| *best/2* | $V_{i,G} = X_{best,G} + F\left(X_{r_1^i,G} - X_{r_2^i,G} + X_{r_3^i,G} - X_{r_4^i,G}\right)$ |
| *current-to-rand/1* | $V_{i,G} = X_{i,G} + K\left(X_{r_3^i,G} - X_{i,G}\right) + F\left(X_{r_1^i,G} - X_{r_2^i,G}\right)$ |
| *current-to-best/1* | $V_{i,G} = X_{i,G} + K\left(X_{best,G} - X_{i,G}\right) + F\left(X_{r_1^i,G} - X_{r_2^i,G}\right)$ |

## 2. RELATED WORKS

Menzura-Montes et. al. [12] empirically compared the performance of eight DE variants on unconstrained optimization problems. Variants with arithmetic recombination, since they are rotationally invariant, were also considered in their work. He used convergence measure to identify the competitiveness of the variants. The study concluded rand/1/bin, best/1/bin, current-to-rand/1/bin and rand/2/dir as the most competitive variants. However, the potential variants like best/2/*, rand-to-best/1/* and rand/2/* were not considered in their study.

Daniela Zaharie [6], provides theoretical insights on explorative power of Differential Evolutional algorithms, she describes an expression as a measure of the explorative power of population-based optimization methods. In her results, she analyzed the evolution of population variance for rand/1/bin for two test functions (Rastrighin and Ackley). Control of diversity and associated parameter tuning are discussed in [7, 8, 9]

Hans-Georg Beyer [10], analyzed how the ES/EP-like algorithms perform the evolutionary search in the real-valued N-dimensional spaces. He described the search behavior as the antagonism of exploitation and exploration, where exploitation works in one dimension, whereas the exploration is a random walk.

## 3. DESIGN OF EXPERIMENTS

In this paper, we identify the competitiveness of DE variants and it is justified by investigating their convergence nature by implementing them on a set of benchmark problems with high dimensionality and different features. We have chosen fourteen test functions [11, 12], of dimensionality 30, grouped by the feature - unimodal separable, unimodal nonseparable, multimodal separable and multimodal nonseparable. The details of the benchmark problems are described in Table 2.

All the test functions have an optimum value at zero except *f08*. In order to show the similar results, the description of *f08* was adjusted to have its optimum value at zero by just adding the optimal value for the function with 30 decision variables (12569.486618164879) [12].

The parameters for all the DE variants were: population size NP = 60 and maximum number of generations = 3000 (consequently, the maximum numbers of function evaluations calculate to 180,000). The moderate population size and number of generations were chosen to demonstrate the efficacy of DE variants in solving the chosen problems. The variants will stop before the maximum number of generations is reached only if the tolerance error (which has been fixed as an error value of 1 x $10^{-12)}$ with respect to the global optimum is obtained. Following [12, 13], we defined a range for the scaling factor, $F \in$ [0.3, 0.9] and this value is generated anew at each generation for all variants. We use the same value for K as F.

Table 2. Details of the test functions used in the experiment

| Functions and Ranges | Functions and Ranges |
|---|---|
| $f_1 : f_{Sp}(x) = \sum_{i=1}^{30} x_i^2$ ; $-100 \leq x_i \leq 100$ | $f_8 : f_{Sch}(x) = \sum_{i=1}^{30}(x_i \sin(\sqrt{|x_i|}))$; $-500 \leq x_i \leq 500$ |
| $f_2 : f_{Sch}(x) = \sum_{i=1}^{30}|x_i| + \prod_{i=1}^{30}|x_i|$ ; $-10 \leq x_i \leq 10$ | $f_9 : f_{Ras}(x) = \sum_{i=1}^{30}[x_i^2 - 10\cos(2\pi x_i) + 10]$; $-5.12 \leq x_i \leq 5.12$ |
| $f_3 : f_{schDS}(x) = \sum_{i=1}^{30}(\sum_{j=1}^{i} x_j)^2$ ; $-100 \leq x_i \leq 100$; | $f_{10} : f_{Ack}(x) =$ $20 + e - 20exp\left(-0.2\sqrt{\frac{1}{p}\sum_{i=2}^{p}x_i^2}\right) - exp\left(\frac{1}{p}\sum_{i=1}^{p}\cos(2\pi x_i)\right)$ ; $-30 \leq x_i \leq 30$ |
| $f_4 : f_{sch}(x) = max_i\{|x_i|, 1 \leq i \leq 30\}$ ; $-100 \leq x_i \leq 100$ | $f_{11} : f_{Gri}(x) = \frac{1}{4000}\sum_{i=1}^{30}x_i^2 - \prod_{i=1}^{30}cos\left(\frac{x_i}{\sqrt{i}}\right) + 1$ ; $-600 \leq x_i \leq 600$ |
| $f_5 : f_{Ros}(x) = \sum_{i=1}^{29}|100(x_{i+1} - x_i^2)^2 + (x_i - 1)^2|$ ; $-30 \leq x_i \leq 30$ | $f_{12} : f_{GPF12} = \frac{\pi}{30}\{10sin^2(\pi y_1) + \sum_{i=1}^{30}(y_i - 1)^2[1 + 10sin^2(\pi y_{i+1})] + (y_n - 1)^2\} + \sum_{i=1}^{30}u(x_i, 10,100,4)$ ; $-50 \leq x_i \leq 50$ |
| $f_6 : f_{St}(x) = \sum_{i=1}^{30}(\lfloor x_i + 0.5\rfloor)^2$ ; $-1.28 \leq x_i \leq 1.28$ | $f_{13} : f_{GPF13} = 0.1\{sin^2(\pi 3x_1) + \sum_{i=1}^{29}(x_i - 1)^2[1 + sin^2(3\pi xy_{i+1})] + (x_n - 1)^2\}[1 + sin^2(2\pi x_{30})] + \sum_{i=1}^{30}u(x_i, 10,100,4)$ ; $-50 \leq x_i \leq 50$ |
| $f_7 : f_{QF}(x) = \sum_{i=1}^{30}ix_i^4 + random[0,1)$ ; $-1.28 \leq x_i \leq 1.28$ | $f_{14} : f_{Boh} = x_i^2 + 2x_{i+1}^2 - 0.3cos(3\pi x_i) - 0.4cos(4\pi x_{i+1}) + 0.7$ ; $-100 \leq x_i \leq 100$ |

The crossover rate, CR, was tuned for each variant-test function combination. Eleven different values for the CR viz. {0.0, 0.1, 0.2, 0.3, 0.4, 0.5, 0.6, 0.7, 0.8, 0.9, 1.0} were tested for each variant-test function combination. For each combination of variant-test function-CR value, 50 independent runs were conducted. Based on the obtained results, a bootstrap test was conducted in order to determine the confidence interval for the mean objective function value. The CR value corresponding to the best confidence interval was chosen to be used in our experiment. The fourteen variants of DE along with the CR values for each test function are presented in Table 3.

As EA's are stochastic in nature, 100 independent runs were performed per variant per test function (by initializing the population for every run with uniform random initialization within the search space). The competitiveness of DE variants in solving the benchmark functions are identified by comparing their mean objective function values (*MOV*). The convergence analysis of the variants are carried out by measuring their Convergence Speed ($C_s$), Quality Measure($Q$-Measure $Q_m$) [14] for each variant-test function combination.

Table 3. "CR" Value for Each Pair of Variant-Function

| Sno | Variant | $f_1/f_8$ | $f_2/f_9$ | $f_3/f_{10}$ | $f_4/f_{11}$ | $f_5/f_{12}$ | $f_6/f_{13}$ | $f_7/f_{14}$ |
|---|---|---|---|---|---|---|---|---|
| 1 | rand/1/bin | 0.9/0.5 | 0.2/0.1 | 0.9/0.9 | 0.5/0.1 | 0.9/0.1 | 0.2/0.1 | 0.8/0.1 |
| 2 | rand/1/exp | 0.9/0.0 | 0.9/0.9 | 0.9/0.9 | 0.9/0.9 | 0.9/0.9 | 0.9/0.9 | 0.9/0.9 |
| 3 | best/1/bin | 0.1/0.1 | 0.1/0.1 | 0.5/0.1 | 0.2/0.1 | 0.8/0.3 | 0.1/0.8 | 0.7/0.1 |
| 4 | best/1/exp | 0.9/0.7 | 0.8/0.9 | 0.9/0.8 | 0.9/0.8 | 0.8/0.9 | 0.8/0.8 | 0.9/0.8 |
| 5 | rand/2/bin | 0.3/0.2 | 0.1/0.1 | 0.9/0.1 | 0.2/0.1 | 0.9/0.1 | 0.2/0.1 | 0.9/0.1 |
| 6 | rand/2/exp | 0.9/0.3 | 0.9/0.9 | 0.9/0.9 | 0.9/0.9 | 0.9/0.9 | 0.9/0.9 | 0.9/0.9 |
| 7 | best/2/bin | 0.1/0.7 | 0.3/0.1 | 0.7/0.4 | 0.2/0.1 | 0.6/0.1 | 0.1/0.1 | 0.5/0.1 |
| 8 | best/2/exp | 0.9/0.3 | 0.9/0.9 | 0.9/0.9 | 0.9/0.9 | 0.9/0.9 | 0.9/0.9 | 0.9/0.9 |
| 9 | current-to-rand/1/bin | 0.5/0.4 | 0.1/0.1 | 0.9/0.1 | 0.2/0.1 | 0.1/0.2 | 0.1/0.3 | 0.2/0.1 |
| 10 | current-to-rand/1/exp | 0.9/0.3 | 0.9/0.9 | 0.9/0.9 | 0.9/0.9 | 0.9/0.9 | 0.9/0.9 | 0.9/0.9 |
| 11 | current-to-best/1/bin | 0.2/0.8 | 0.1/0.1 | 0.9/0.1 | 0.2/0.2 | 0.1/0.2 | 0.3/0.1 | 0.2/0.1 |
| 12 | current-to-best/1/exp | 0.9/0.1 | 0.9/0.9 | 0.9/0.9 | 0.9/0.9 | 0.9/0.9 | 0.9/0.9 | 0.9/0.9 |
| 13 | rand-to-best/1/bin | 0.1/0.8 | 0.1/0.1 | 0.9/0.9 | 0.4/0.1 | 0.8/0.1 | 0.4/0.2 | 0.8/0.1 |
| 14 | rand-to-best/1/exp | 0.9/0.4 | 0.9/0.9 | 0.9/0.9 | 0.9/0.9 | 0.9/0.9 | 0.9/0.9 | 0.9/0.9 |

## 3. EMPIRICAL MEASUREMENTS

Convergence Speed is used to detect which variant is most competitive. To measure the convergence speed, we calculated the mean percentage out of the total 1, 80,000 function evaluations required by each of the variant to reach its best objective function value, for all the 100 independent runs

The algorithm convergence, usually mean the convergence of the objective function that we minimize. The rate of convergence is generally described by $f(gen)$, where $gen$ is the current generation. We use $Q$-Measure to study the convergence nature of DE Variants. Quality measure or simply $Q$-Measure is an empirical measure of the algorithm's convergence. It serves to compare the objective function convergence of different evolutionary methods. In our experiment, it is used to study the behavior of our DE variants.

The formula of $Q$-Measure is $Q_m = C / P_c$
$P_c$ – Probability of convergence,
$C$ – Convergence Measure.

The Convergence Measure ($C$) is calculated as $C = SumEj / nc$
$nc$ – number of successful runs
$SumEj$ – total number of function evaluations taken for all the successful runs

## 4. RESULTS AND DISCUSSION

The mean objective function values obtained for the unimodal separable functions: $f_1$, $f_2$, $f_4$, $f_6$ and $f_7$, and the unimodal nonseparable function $f_3$ are presented in Table 4. The results shows that best performance were provided by rand-to-best/1/bin, rand/1/bin, best/2/bin and

*rand/2/bin* variants for the unimodal separable functions. *best/1/\*, current-to-rand/1/exp* and *current-to-best/1/exp* were the poorly performing variants. *best/2/\** variants alone was able to solve the unimodal nonseparable problem. The variant *current-to-rand/1/\** and *current-to-best/1/\** displayed the worst performance.

Table 5 displays the simulation results for the multimodal separable functions: $f_8$, $f_9$ and $f_{14}$, and for the multimodal non-separable functions $f_5$, $f_{10}$, $f_{11}$, $f_{12}$ and $f_{13}$. In case of multimodal separable problems, the best performance was shown by *rand/1/bin, rand-to-best/1/bin* and *rand/2/bin* variants once again as in the case of unimodal separable problems. Similarly, the variants current-to-rand/1/exp and *current-to-best/1/exp* consistently showed poor performance. In the case of multimodal nonseparable problems, function $f_5$ and $f_{10}$ were not solved by any variants.

In this case of $f_{11}$, $f_{12}$ and $f_{13}$, the best performing variants were *rand-to-best/1/bin, rand/1/bin, rand/2/bin* and *best/2/bin*, once again along with *current-to-rand/1/bin* and *rand-to-best/1/bin*. *current-to-rand/1/exp* and *current-to-best/1/exp* were the poorly performing variants along with *best/1/\** variants.

Table 4. MOV Obtained for Unimodal functions

| Variant | $f_1$ | $f_2$ | $f_4$ | $f_6$ | $f_7$ | $f_3$ |
|---|---|---|---|---|---|---|
| *rand/1/bin* | 0.00 | 0.00 | 0.00 | 0.02 | 0.00 | 0.07 |
| *rand/1/exp* | 0.00 | 0.00 | 3.76 | 0.00 | 0.02 | 0.31 |
| *best/1/bin* | 457.25 | 0.14 | 1.96 | 437.25 | 0.09 | 13.27 |
| *best/1/exp* | 583.79 | 4.05 | 37.36 | 591.85 | 0.06 | 57.39 |
| *rand/2/bin* | 0.00 | 0.00 | 0.06 | 0.00 | 0.01 | 1.64 |
| *rand/2/exp* | 0.00 | 0.02 | 32.90 | 0.00 | 0.05 | 269.86 |
| *best/2/bin* | 0.00 | 0.00 | 0.00 | 0.07 | 0.00 | 0.00 |
| *best/2/exp* | 0.00 | 0.00 | 0.05 | 0.39 | 0.01 | 0.00 |
| *current-to-rand/1/bin* | 0.00 | 0.02 | 3.68 | 0.03 | 0.04 | 3210.36 |
| *current-to-rand/1/exp* | 24.29 | 44.22 | 57.52 | 43.07 | 0.27 | 3110.90 |
| *current-to-best/1/bin* | 0.00 | 0.02 | 3.71 | 0.00 | 0.04 | 3444.00 |
| *current-to-best/1/exp* | 24.37 | 45.04 | 56.67 | 41.95 | 0.26 | 2972.62 |
| *rand-to-best/1/bin* | 0.00 | 0.00 | 0.00 | 0.00 | 0.00 | 0.07 |
| *rand-to-best/1/exp* | 0.00 | 0.00 | 3.38 | 0.00 | 0.01 | 0.20 |

Based on the overall results in Table 4 and 5 the most competitive variants were *rand-to-best/1/bin, best/2/bin* and *rand/1/bin*. The variants *rand/2/bin* and *best/2/exp* also showed good performance consistently. On the other hand, the worst overall performances were consistently displayed by variants *current-to-best/1/exp* and *current-to-rand/1/exp*. The variants *best/1/exp* and *current-to-rand/bin* were also displaying poor performance. *best/2/\** variants show good performance for unimodal nonseparable functions. It is worth noting that binomial recombination showed a better performance over the exponential recombination.

Table 5 : MOV obtained for Multimodal functions

| Variant | $f_8$ | $f_9$ | $f_{14}$ | $f_5$ | $f_{10}$ | $f_{11}$ | $f_{12}$ | $f_{13}$ |
|---|---|---|---|---|---|---|---|---|
| rand/1/bin | 0.13 | 0.00 | 0.00 | 21.99 | 0.09 | 0.00 | 0.00 | 0.00 |
| rand/1/exp | 0.10 | 47.93 | 0.00 | 25.48 | 0.09 | 0.05 | 0.00 | 0.00 |
| best/1/bin | 0.00 | 4.33 | 12.93 | 585899.88 | 3.58 | 3.72 | 15.78 | 973097.03 |
| best/1/exp | 0.01 | 50.74 | 32.18 | 64543.84 | 6.09 | 5.91 | 131448.66 | 154434.94 |
| rand/2/bin | 0.22 | 0.00 | 0.00 | 19.01 | 0.09 | 0.00 | 0.00 | 0.00 |
| rand/2/exp | 0.27 | 101.38 | 0.01 | 2741.32 | 0.01 | 0.21 | 0.00 | 0.01 |
| best/2/bin | 0.17 | 0.69 | 0.12 | 2.32 | 0.09 | 0.00 | 0.00 | 0.00 |
| best/2/exp | 0.08 | 80.63 | 2.53 | 1.12 | 0.83 | 0.03 | 0.14 | 0.00 |
| current-to-rand/1/bin | 0.14 | 37.75 | 0.00 | 52.81 | 0.01 | 0.00 | 0.00 | 0.00 |
| current-to-rand/1/exp | 0.12 | 235.14 | 18.35 | 199243.32 | 13.83 | 1.21 | 10.89 | 24.11 |
| current-to-best/1/bin | 0.19 | 37.04 | 0.00 | 56.91 | 0.01 | 0.00 | 0.00 | 0.00 |
| current-to-best/1/exp | 0.10 | 232.80 | 18.21 | 119685.68 | 13.69 | 1.21 | 10.37 | 23.04 |
| rand-to-best/1/bin | 0.22 | 0.00 | 0.00 | 17.37 | 0.09 | 0.00 | 0.00 | 0.00 |
| rand-to-best/1/exp | 0.12 | 48.09 | 0.00 | 24.54 | 0.09 | 0.05 | 0.00 | 0.00 |

Next in our experiment, we measure the Convergence Speed ($C_s$) as the mean percentage of total number of function evaluations required by each of the variant, for 100 runs, to reach their best objective function value. This measure is used to detect which variants are more competitive, in each set of function classes. The variants which are good in convergence will have less mean percentage ie, they will reach their optimum value with less number of function evaluations. The result in Table 6 shows that the variants *rand/1/bin, rand/2/bin, best/2/bin, rand-to-best/1/bin* and *best/2/exp* are faster in convergence to global optimum. The variants which are taking 100% of the function evaluations, ie., poor performance, for all the functions are *best/1/bin, best/1/exp, current-to-rand/1/exp* and *current-to-best/1/exp*. The variants which are providing the percentage less than 100, for only one function, are *rand/1/exp, rand/2/exp, current-to-rand/1/bin, current-to-best/1/bin* and *rand-to-best/1/exp*. The results suggest that the top four variants which could reach the best objective function values with less number of function evaluations are *rand/1/bin, rand-to-best/1/bin, best/2/bin* and r*and/2/bin*. The variants which perform poor are *best/1/bin, best/1/exp, current-to-rand/1/exp* and *current-to-best/1/exp*.

We analyzed the performance of the variants by each function class wise. We observed that for the unimodal functions the top four variants with the fastest convergence are *rand/1/bin, rand/2/bin, best/2/bin* and *rand-to-best/1/bin*. For $f_4$ and $f_7$ all the variants behaves similar in their speed of convergence, but *rand/1/bin, best/2/bin* and *rand-to-best/1/bin* could reach the global optimum. The results suggest that, the variant *rand/1/bin* out performing others by its convergence speed and the variants *best/1/bin, best/1/exp, current-to-rand/1/exp* and *current-to-best/1/exp* are slow in convergence. For $f_3$, *best/2/exp* is comparatively faster than other variants by 0.05%. For the multimodal functions $f_5, f_8, f_9, f_{10}, f_{11}, f_{12}, f_{13}$ and $f_{14}$ the top four variants with the fastest convergence are *rand-to-best/1/bin, rand/1/bin, best/2/bin* and *rand/2/bin* among these variants *rand-to-best/1/bin* and *rand/1/bin* are equally faster in convergence than others. For the function $f_8$ and $f_{10}$, all the variants behave similar in their speed of convergence. The results suggest that, the variants *rand-to-best/1/bin* and *rand/1/bin* are out performing others by its convergence speed and the variants which week or slow in converging to the optimum value are *best/1/bin, best/1/exp, current-to-rand/1/exp* and *current-to-best/1/exp*.

Table 6. Convergence Speed Measured for the Variants. The lowest percentage is marked with "*"

| Variant | $f_1$ | $f_2$ | $f_3$ | $f_4$ | $f_5$ | $f_6$ | $f_7$ | $f_8$ | $f_9$ | $f_{10}$ | $f_{11}$ | $f_{12}$ | $f_{13}$ | $f_{14}$ |
|---|---|---|---|---|---|---|---|---|---|---|---|---|---|---|
| rand/1/bin | 40.93* | 56.45 | 100 | 100 | 100 | 10.89* | 100 | 100 | 65.50 | 100 | 46.89 | 38.89* | 41.38 | 42.13 |
| rand/1/exp | 100 | 100 | 100 | 100 | 100 | 39.34 | 100 | 100 | 100 | 100 | 100 | 100 | 100 | 100 |
| best/1/bin | 100 | 100 | 100 | 100 | 100 | 100 | 100 | 100 | 100 | 100 | 100 | 100 | 100 | 100 |
| best/1/exp | 100 | 100 | 100 | 100 | 100 | 100 | 100 | 100 | 100 | 100 | 100 | 100 | 100 | 100 |
| rand/2/bin | 70.05 | 90.57 | 100 | 100 | 100 | 17.78 | 100 | 100 | 100 | 100 | 70.02 | 56.07 | 58.62 | 59.85 |
| rand/2/exp | 100 | 100 | 100 | 100 | 100 | 81.64 | 100 | 100 | 100 | 100 | 100 | 100 | 100 | 100 |
| best/2/bin | 42.67 | 48.9* | 100 | 100 | 99.97* | 12.47 | 100 | 100 | 87.29 | 100 | 49.60 | 40.05 | 41.23 | 48.60 |
| best/2/exp | 73.42 | 100 | 99.85* | 100 | 100 | 51.93 | 100 | 100 | 100 | 100 | 88.94 | 80.43 | 81.42 | 98.35 |
| current-to-rand/1/bin | 100 | 100 | 100 | 100 | 100 | 40.13 | 100 | 100 | 100 | 100 | 100 | 100 | 100 | 100 |
| current-to-rand/1/exp | 100 | 100 | 100 | 100 | 100 | 100 | 100 | 100 | 100 | 100 | 100 | 100 | 100 | 100 |
| current-to-best/1/bin | 100 | 100 | 100 | 100 | 100 | 50.84 | 100 | 100 | 100 | 100 | 100 | 100 | 100 | 100 |
| current-to-best/1/exp | 100 | 100 | 100 | 100 | 100 | 100 | 100 | 100 | 100 | 100 | 100 | 100 | 100 | 100 |
| rand-to-best/1/bin | 43.11 | 63.57 | 100 | 100 | 100 | 11.45 | 100 | 100 | 65.10* | 100 | 46.60* | 38.98 | 38.34* | 42.09* |
| rand-to-best/1/exp | 100 | 100 | 100 | 100 | 100 | 39.11 | 100 | 100 | 100 | 100 | 100 | 100 | 100 | 100 |

Next to Convergence Speed, we measured *Q-Measure*. This measure provides a more objective vision of the behavior of a DE variant. It is an integral measure that combines the convergence of a DE algorithm with its probability of convergence. By combining the convergence rate and its probability in one, now the $Q_m$ is a single criterion to be minimized. From the $Q_m$ values measured for all the variant-function combinations, we identified that the variants with least $Q_m$ value are *rand-to-best/1/bin, best/2/bin, rand/1/bin* and *rand/2/bin*. The variants with higher $Q_m$ value, ie., poor in performance, are *best/1/bin, best/1/exp, current-to-best/1/exp* and *current-to-rand/1/exp*. $Q_m$ value increases as probability of convergence decreases, ie., variants with lower probability of convergence gives higher value for quality measure and vice versa. The results suggest that the variants with good convergence rate and higher probability of convergence gives minimum value of $Q_m$, such variants are *rand-to-best/1/bin, best/2/bin, rand/1/bin and rand/2/bin*. The variants *best/1/bin, best/1/exp, current-to-rand/1/exp and current-to-best/1/exp* are slow in convergence.

For the unimodal functions, the values of $Q_m$ are presented in the Table 7. The results show that the top four variants with least $Q_m$ values are *rand/1/bin, best/2/bin, rand-to-best/1/bin* and *rand/2/bin*. The variants *best/1/exp, current-to-rand/1/exp* and *current-to-best/1/exp* could not provide any successful run. The variant with highest $Q_m$ value, id., poorest performance, is *best/1/bin*. The results suggest that, even though some of the "bin" variants behave worst, all the variants in the top are "bin" variants. Similarly, even though some of the "exp" variants could give moderate $Q_m$ value, all the variants which could not provide any successful run are "exp" variants. This shows the influence of binomial recombination in DE variants.

For the function $f_3$, results (Table 8) show that the top four variants with less $Q_m$ values are *best/2/bin, best/2/exp, best/1/bin* and *rand-to-best/1/bin*. The variant *rand/1/bin* is next in the list. The number of successful runs made by the variants *current-to-bes/1/exp, rand/2/exp, rand/1/bin*,

*current-to-rand/1/bin*, *current-to-rand/1/exp*, *current-to-best/1/bin* are zero. For $f_3$ , the variants with higher $Q_m$ values, ie., poor performance, are *best/1/exp* , *rand-to-best/1/exp* and *rand/1/exp*. The convergence speeds of all the variants are equal, because all the variants have reached the global optimum at *MaxFE* only.

Table 7. *Q-Measure* achieved by the variants for unimodal separable functions, in the ascending order of *Q-Measure*.

| Variant | $f_1$ | $f_2$ | $f_4$ | $f_6$ | $f_7$ | *SumEj* | *C* | $Q_m=C/P_c$ |
|---|---|---|---|---|---|---|---|---|
| *rand/1/bin* | 73686.6 | 1E+07 | 18000000 | 1922580 | 10800000 | 40958347 | 89428.70 | 976.30 |
| *best/2/bin* | 7679760 | 8819160 | 18000000 | 2135280 | 13500000 | 50134200 | 106668.51 | 1134.77 |
| *rand-to-best/1/bin* | 7759500 | 1.1E+07 | 18000000 | 2061060 | 10800000 | 50062560 | 108831.65 | 1182.95 |
| *rand/2/bin* | 1.3E+07 | 1.6E+07 | - | 3201060 | 360000 | 32474740 | 107532.25 | 1780.34 |
| *rand/1/exp* | 1.8E+07 | 1.8E+07 | - | 7081860 | - | 43081860 | 143606.20 | 2393.44 |
| *best/2/exp* | 1.3E+07 | 1.8E+07 | 180000 | 3767640 | 360000 | 35522400 | 130597.06 | 2400.68 |
| *rand-to-best/1/exp* | 1.8E+07 | - | - | 7039200 | - | 25039200 | 125196.00 | 3129.90 |
| *current-to-rand/1/bin* | 1.8E+07 | - | - | 7222800 | - | 25222800 | 126114.00 | 3152.85 |
| *current-to-best/1/bin* | 1.8E+07 | - | - | 9150660 | - | 27150660 | 135753.30 | 3393.83 |
| *rand/2/exp* | 1.1E+07 | - | - | 14695800 | - | 25675800 | 159477.02 | 4952.70 |
| *best/1/bin* | 540000 | 7200000 | 14220000 | 0 | 3960000 | 25920000 | 180000.00 | 6250.00 |
| *best/1/exp* | - | - | - | - | - | - | - | - |
| *current-to-rand/1/exp* | - | - | - | - | - | - | - | - |
| *current-to-best/1/exp* | - | - | - | - | - | - | - | - |

Table 8. *Q-Measure* achieved by the variants for unimodal nonseparable, in the ascending order of *Q-Measure*.

| Variant | $f_3$ | *SumEj* | *C* | $Q_m=C/P_c$ |
|---|---|---|---|---|
| *best/2/bin* | 18000000 | 18000000 | 180000 | 1800 |
| *best/2/exp* | 18000000 | 18000000 | 180000 | 1800 |
| *best/1/bin* | 15480000 | 15480000 | 180000 | 2093.02 |
| *rand-to-best/1/bin* | 14220000 | 14220000 | 180000 | 2278.48 |
| *rand/1/bin* | 13140000 | 13140000 | 180000 | 2465.75 |
| *best/1/exp* | 10440000 | 10440000 | 180000 | 3103.45 |
| *rand-to-best/1/exp* | 1800000 | 1800000 | 180000 | 18000 |
| *rand/1/exp* | 720000 | 720000 | 180000 | 45000 |
| *current-to-best/1/exp* | - | - | - | - |
| *rand/2/exp* | - | - | - | - |
| *rand/2/bin* | - | - | - | - |
| *current-to-rand/1/bin* | - | - | - | - |
| *current-to-rand/1/exp* | - | - | - | - |
| *current-to-best/1/bin* | - | - | - | - |

The $Q_m$ values measured for the multimodal functions are presented in Table 9 and 10. The results show that the variants which are in the top with least $Q_m$ values are *rand-to-best/1/bin, rand/1/bin, best/2/bin* and *rand/2/bin*. The variants which are in the bottom of the list with higher $Q_m$ values are *rand/2/exp*, *best/2/exp*, *current-to-best/1/exp* and *current-to-rand/1/exp*. For the multimodal separable functions, the "bin" variants are outperforming the "exp" variants with good convergence rate and higher probability of convergence. For the multimodal nonseparable functions, the variant with higher values of $Q_m$ is *best/1/bin*. For the variants *current-to-rand/1/exp*, *current-to-best/1/exp* and *best/1/exp* the number of successful runs are zero. The results suggest the influence of binomial recombination, the variants with higher convergence rate and higher probability of convergence are "*bin*" variants. But due to its less probability of convergence (0.2%), the variant *best/1/bin* could not perform well as other "*bin*" variants.

Table 9. Q-measure achieved by the variants for multimodal separable functions, in the ascending order of Q-Measure.

| Variant | $f_8$ | $f_9$ | $f_{14}$ | SumEj | C | $Q_m=C/P_c$ |
|---|---|---|---|---|---|---|
| *rand-to-best/1/bin* | - | 1.2E+07 | 7576620 | 1.9E+07 | 96476.1 | 1447.14 |
| *rand/1/bin* | 720000 | 1.2E+07 | 7584180 | 2E+07 | 98504.1 | 1448.59 |
| *best/2/bin* | 180000 | 6171360 | 6767700 | 1.3E+07 | 95759.6 | 2096.92 |
| *rand/2/bin* | 180000 | 1.8E+07 | 1.1E+07 | 2.9E+07 | 144040 | 2149.85 |
| *rand/1/exp* | 1260000 | - | 1.8E+07 | 1.9E+07 | 180000 | 5046.73 |
| *rand-to-best/1/exp* | 1080000 | - | 1.8E+07 | 1.9E+07 | 180000 | 5094.34 |
| *current-to-best/1/bin* | 540000 | - | 1.8E+07 | 1.9E+07 | 180000 | 5242.72 |
| *current-to-rand/1/bin* | 360000 | - | 1.8E+07 | 1.8E+07 | 180000 | 5294.12 |
| *best/1/bin* | 1.6E+07 | 540000 | - | 1.6E+07 | 180000 | 5934.07 |
| *best/1/exp* | 1.5E+07 | - | - | 1.5E+07 | 180000 | 6352.94 |
| *rand/2/exp* | 360000 | - | 4680000 | 5040000 | 180000 | 19285.7 |
| *best/2/exp* | 3060000 | - | 956880 | 4016880 | 167370 | 20921.3 |
| *current-to-best/1/exp* | 900000 | - | - | 900000 | 180000 | 108000 |
| *current-to-rand/1/exp* | 540000 | - | - | 540000 | 180000 | 180000 |

Table 10. *Q-Measure* achieved by the variants for multimodal nonseparable functions, in the ascending order of *Q-Measure*

| Variant | $f_5$ | $f_{10}$ | $f_{11}$ | $f_{12}$ | $f_{13}$ | SumEj | C | $Q_m=C/P_c$ |
|---|---|---|---|---|---|---|---|---|
| *rand-to-best/1/bin* | - | - | 8388600 | 7008720 | 6901200 | 22298520 | 74328.40 | 1238.81 |
| *rand/1/bin* | - | - | 8437680 | 7000500 | 7448880 | 22887060 | 76290.20 | 1271.50 |
| *best/2/bin* | 6835320 | - | 8927340 | 7028940 | 7422060 | 30213660 | 89654.78 | 1330.19 |
| *rand/2/bin* | - | - | 12603420 | 10092120 | 10551840 | 33247380 | 110824.6 | 1847.08 |
| *current-to-rand/1/bin* | - | 1E+07 | 17280000 | 17280000 | 18000000 | 62640000 | 180000 | 2586.21 |
| *current-to-best/1/bin* | - | 1E+07 | 17280000 | 16380000 | 18000000 | 61740000 | 180000 | 2623.91 |
| *best/2/exp* | 5220000 | - | 5929380 | 8896560 | 9076380 | 29122320 | 138020.47 | 3270.63 |
| *rand-to-best/1/exp* | - | - | 12420000 | 18000000 | 18000000 | 48420000 | 180000 | 3345.72 |
| *rand/1/exp* | - | - | 12240000 | 18000000 | 18000000 | 48240000 | 180000 | 3358.21 |
| *rand/2/exp* | - | 1.2E+07 | 540000 | 18000000 | 9000000 | 39600000 | 180000 | 4147.47 |
| *best/1/bin* | - | - | 180000 | - | - | 180000 | 180000 | 900000.00 |
| *current-to-rand/1/exp* | - | - | - | - | - | - | - | - |
| *current-to-best/1/exp* | - | - | - | - | - | - | - | - |
| *best/1/exp* | - | - | - | - | - | - | - | - |

## 5. CONCLUSION

In this paper, the competiveness and convergence nature of different variants of Differential Evolution algorithm are analyzed. Empirical comparison of fourteen DE variants to solve fourteen global optimization problems is done. The competiveness of the variants is analyzed by their mean objective function values and the best and worst performing variants are identified. Regardless of the characteristics and dimension of the functions, relatively better results seem to have been provided by the variants with binomial crossover. The competitiveness of the variants is validated by analyzing their convergence behavior, by measuring their convergence speed and quality measure. The results suggest that the best performing variants (*rand/1/bin*, *rand/2/bin*, *best/2/bin*, *rand-to-best/1/bin* and *best/2/exp)* are faster in converging to the solution. The worst performing variants were found to have less probability of convergence, and hence they were slow in convergence. The identified difference in convergence behavior of the variants may be due to either stagnation problem or premature convergence, which are lead by improper balance between exploration and exploitation processes. This work can be still further analyzed by focusing on improving the performance of variants in the light of bringing balance between explorations exploitation during the generations.

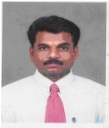

**G.Jeyakumar** received his B.Sc degree in Mathematics in 1994 and his M.C.A degree (under the faculty of Engineering) in 1998 from Bharathidasan University, Tamil Nadu, India. He is currently an Assistant Professor(Selection Grade) in the department of Computer Science and Engineering, Amrita School of Engineering , Amrita Vishwa Vidyapeetham University, Tamil Nadu, India since 2000. His research interest include evolutionary algorithm, differential evolution, parallelization of differential evolutions and applications of differential evolution.

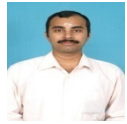

**C. Shunmuga Velayutham** received the B.Sc degree in Physics from Manonmaniam Sundaranar University, Tamilnadu, India, in 1998, M.Sc degree in Electronics and Ph.D. degree in Neuro-Fuzzy Systems from Dayalbagh Educational Institute, Uttar Pradesh, India in 2000 and 2005 respectively. Currently, he is an Assistant Professor in the Department of Computer Science & Engineering, Amrita Vishwa Vidyapeetham, Tamilnadu, India since 2005. His research encompasses theoretical investigation and application potential (esp. in computer vision) of evolutionary computation.